\documentclass[]{spie}  

\usepackage{amsmath,amsfonts,amssymb}
\usepackage{graphicx}
\usepackage[colorlinks=true, allcolors=blue]{hyperref}

\usepackage{float} 

\title{Building Brain Tumor Segmentation Networks with User-Assisted Filter Estimation and Selection}

\author[a]{Matheus A. Cerqueira}
\author[b]{Flávia Sprenger}
\author[b,c]{Bernardo C. A. Teixeira}
\author[a]{Alexandre X. Falcão}
\affil[a]{Institute of Computing, University of Campinas, Campinas, São Paulo, Brazil}
\affil[b]{Hospital de Clínicas,  Universidade Federal do Paraná, Curitiba, Paraná, Brazil}
\affil[c]{Instituto de Neurologia de Curitiba, Curitiba, Paraná, Brazil}

\authorinfo{This is a manuscript,further author information: (Send correspondence to M.A.C.)\\M.A.C.: E-mail: matheus.cerqueira@students.ic.unicamp.br\\  F.S.: E-mail: flaviasprenger@gmail.com\\ B.C.A.T.: E-mail: berteixeira@gmail.com\\ A.X.F.: E-mail: afalcao@ic.unicamp.br}

\begin{document} 
\maketitle

\begin{abstract}
\textit{Brain tumor image segmentation} is a challenging research topic in which deep-learning models have presented the best results. However, the traditional way of training those models from many pre-annotated images leaves several questions unanswered. Hence methodologies, such as \textit{Feature Learning from Image Markers} (FLIM), have involved an expert in the learning loop to reduce human effort in data annotation and build models sufficiently deep for a given problem. FLIM has been successfully used to create encoders, estimating the filters of all convolutional layers from patches centered at marker voxels. In this work, we present Multi-Step (MS) FLIM -- a user-assisted approach to estimating and selecting the most relevant filters from multiple FLIM executions. MS-FLIM is used only for the first convolutional layer, and the results already indicate improvement over FLIM. For evaluation, we build a simple U-shaped encoder-decoder network, named \textit{sU-Net}, for glioblastoma segmentation using T1Gd and FLAIR MRI scans, varying the encoder's training method, using FLIM, MS-FLIM, and backpropagation algorithm. Also, we compared these sU-Nets with two State-Of-The-Art (SOTA) deep-learning models using two datasets. The results show that the sU-Net based on MS-FLIM outperforms the other training methods and achieved effectiveness within the standard deviations of the SOTA models.

\end{abstract}


\section{Introduction}
\label{sec:intro}  


Gliomas are the most common type of brain tumor in adults, with the highest occurrence in brain regions, although other areas of the Central Nervous System (CNS) might be affected~\cite{ostrom2015cbtrus}. They can be separated into Low-Grade (LGG)  and High-Grade (HGG) gliomas, being glioblastoma (GBM) part of the last group. GBM is the most common malignant brain tumor of the CNS in humans – its treatment is complicated, with only a 5\% chance of survival within five years after diagnosis. In both LGGs and HGGs, the initial diagnosis is performed through Computerized Tomography (CT) or Magnetic Resonance Imaging (MRI), being MRI the best option due to multiple scanning protocols that target specific areas of the tumor. Additionally to the diagnosis, volume estimation is essential for monitoring and studying tumor progression and analyzing the adopted treatment~\cite{simi2015segmentation,dupont2016image}. However, manual annotation is time-consuming, tedious, and error-prone -- facts that have motivated research on automatic and semi-automatic methods for brain tumor segmentation.

BRATS (Brain Tumor Image Segmentation) challenge is an example of the efforts to improve brain tumor image analysis by promoting and revealing the best machine learning models~\cite{menze2014multimodal, bakas2017advancing, bakas2018identifying}. This segmentation challenge is made of MRI images of LGG and HGG gliomas, where each sample has 4 MRI scans (T1, T2, T1Gd, and FLAIR), each manually annotated by a group of professionals with years of experience. The dataset also contains the division of training, validation, and test sets, with training and validation sets available to download. The test set is only used for the final evaluation of the challenge. Furthermore, the training set contains the scans and segmentation mask; However, the validation set contains only the input images requiring the submission at the online platform for obtaining the performance metrics. Due to its easy access and relatively high number of annotated images (in 2021, there were 1,251 samples in the training set), the BRATS dataset is used in several works, even for those who never participated in the segmentation challenge.

Nowadays, among the automatic segmentation techniques, the ones based on deep neural networks present the best results for brain tumor segmentation~\cite{wadhwa2019review}. However, the traditional way of training those networks from many pre-annotated images leaves several questions unanswered: How many training images are required for human annotation? What is the network architecture that offers the best cost-benefit for our problem? Is it possible to build such a network layer by layer, selecting the most relevant filters? Such issues require an expert in the machine learning loop. A recent technique~\cite{de2020learning}, named Feature Learning from Image Markers (FLIM), addresses some of these questions by allowing the designer to draw scribbles (markers) in image regions that are expected to be activated by the network filters. The filters of all convolutional layers in the feature extractor (encoder) are estimated from those marker pixels with no backpropagation and using very few training images. Afterward, the encoder can be used with a classifier/decoder, and the whole network can be trained by backpropagation with annotated images, further improving the encoder.

This work proposes a new method, \textit{MultiStep FLIM} (MS-FLIM), in which an encoder is built layer by layer with the help of an expert in the learning loop, assuring that the selected filters activate all regions related to the problem. For this work, the user only participates in filter estimation and selection for the first layer. The user draws markers on all regions of interest in a few T1Gd and FLAIR images. For the first layer, FLIM executes several times, estimating different numbers of filters per marker and image. The user observes the filters' activations and selects the most representative ones for the first layer. FLIM automatically estimates the filters of the remaining layers. To evaluate the impact of user intervention, we build a U-shaped encoder-decoder network named \textit{sU-Net} and compare the performances of the sU-Net models trained with different techniques: FLIM, MS-FLIM and only backpropagation. We used two SOTA deep-learning models as baselines and two datasets with T1Gd and FLAIR scans. The decoder of all sU-Net models is trained with backpropagation, and, during this process, we also evaluate the possibility of fine-tuning the MS-FLIM encoder.

\section{Related work}
\label{sec:related_work}

Related works involve fully and weakly supervised CNN-based models. FLIM may fall in the category of very weakly supervised models, although the version used here relies on unsupervised techniques to estimate filters from user-drawn scribbles (markers). 

\subsection{CNN-based models}
\label{sec:rw_cnn}

Among CNN-based frameworks for brain tumor segmentation, two representative ones are U-Net and DeepMedic. The former is a symmetrical U-shaped encoder-decoder network with convolutional blocks that learns features from different scales. The U-Net's encoder reduces the image size by stridden pooling operations after each convolutional block, using skip connections from the encoder to the decoder before each pooling~\cite{ronneberger2015u}. DeepMedic is a multi-pathway network, where each pathway extracts features from a different input-image resolution, using reshaping and concatenation in the end. In de decoder, the last convolutional layer of both models uses $c$ filters of shape $1\times 1\times 1 \times N$, being $N$ the number of input channels and $c$ the number of classes~\cite{kamnitsas2017efficient, kamnitsas2016deepmedic}.

Nowadays, the best results for brain tumor segmentation depend on very complex deep-learning models based on multiple instances of the U-Net~\cite{jiang2019two} and DeepMedic~\cite{kamnitsas2017efficient, kamnitsas2016deepmedic} networks, or with additional  branches to those models. For example, they increase complexity by adding an autoencoder regularization branch~\cite{myronenko20183d}. Numerous works use patches or slices as network input~\cite{zhong20202wm, jungo2017towards,wang2017automatic} instead of using the entire 3D image, revealing their high computational requirement. Besides, nnU-Net is the best example of customization of those two representative models since it adapts the architecture to different datasets and tasks by adding modifications to the U-Net model. For example, it uses batch normalization and a variety of data augmentation operations~\cite{isensee2021nnu_nature}. The first nnU-Net model was the best in the BRATS 2020 segmentation challenge~\cite{isensee2020nnu_brain} and a variant of it won the 2021 challenge.~\footnote{\url{https://www.rsna.org/education/ai-resources-and-training/ai-image-challenge/brain-tumor-ai-challenge-2021.} }

\subsection{Weakly supervised models}
\label{sec:rw_weakly}
Some works have used partially (weakly or sparsely) annotated images instead of fully-annotated (or densely annotated) images or a combination of both to reduce human effort in data annotation. In~\cite{mlynarski2019deep}, the authors use both annotation types and a U-Net model with a classification branch at the last decoder's layer. This classifier indicates only the presence or absence of the tumor in each slice. The model does not surpass standard deep-learning models trained with many fully-annotated images; However, it can be better than the U-Net model when the number of fully-annotated images is low (e.g., 5-15 images).

Coarse labels can also be generated from user-drawn scribbles (markers). In~\cite{ji2019scribble}, the authors use scribbles and a region-growing method to generate rough labels to train a modified U-Net. They evaluate the technique with prostate cancer and heart segmentation datasets. In the brain tumor segmentation context,  coarse labels have been used to train a first U-Net model for segmenting whole brain lesions. In this case, the multiple regions of interest (classes) inside a lesion are estimated by clustering and used to train a second sub-region U-Net~\cite{ji2019scribble}. This pipeline achieves good results for the complete lesion segmentation but performs poorly for the multi-label segmentation, even with two U-Nets models and using loss based on conditional random field to refine boundaries.


\subsection{The FLIM methodology}
\label{sec:rw_flim}

Rather than propagating labels from scribbles for image annotation, FLIM can estimate the filters of all convolutional layers of an encoder directly from the scribbles. 

The user selects a few images and draws scribbles (markers) in representative regions of interest to the problem. For a given layer, 3D patches are extracted from the centers of all scribble voxels using the shape of the filters and forming a patch dataset. This dataset is centralized similarly to batch normalization but only uses the scribble voxels' region information. One may cluster the patches into $n$ groups and use their centers to compose $n$ desired filters with unit norms to avoid preferences among them. Understanding this process requires considering the convolution between a filter and an image as the dot product between the filter's weight vector and the patch's vectorization at each voxel. The filter's weight vector represents an orthogonal vector to a hyperplane in the local feature space defined by patch extraction. Image patches (e.g., the ones in the same filter's cluster) on the positive side of this hyperplane will be activated. Therefore, the selected filter bank is expected to highlight all regions of interest to the problem.

Since the number of distinct clusters may be much higher than the number of desired filters in a convolutional layer, one can force a given number of clusters per marker and then reduce the candidate filters (cluster centers) by Principal Component Analysis (PCA). This strategy explores the patch distribution in the local feature space by selecting their eigenvectors as the desired filters. Besides activation, the layer can also include other operations (e.g., pooling and skip connections).

The filters of each next layer are then estimated similarly using the scribbles projected onto the previous layer's output. Finally, depending on the application, one can connect the FLIM encoder's output (or one output per layer) to other components. For instance, it can be connected to a multi-layer perceptron for image classification~\cite{de2020learning}, a delineation algorithm for image segmentation~\cite{de2020feature}, or a decoder for regression problems -- the option presented for image segmentation in this work.

\section{The proposed MS-FLIM}
\label{sec:methodolody}

To guarantee relevant filters in the initial convolutional layers, activating all regions of interest to the problem, the proposed MS-FLIM only requires user intervention for filter estimation and selection in the first convolutional layer. FLIM estimates the filters of the remaining layers from the user-drawn markers, as described above. 

\begin{figure}[H]
  \begin{center}
    \begin{tabular}{cc} 
      \includegraphics[scale=0.7]{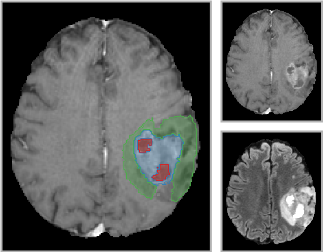}\\
    \end{tabular}
  \end{center}
  \caption{A GBM tumor ground-truth segmentation showing the regions of interest (left): ED (green), ET (blue), and NC (red). ED appears with saturated and intermediate values (right).}
  \label{fig:regions}
\end{figure}

\begin{figure}[ht]
  \begin{center}
    \begin{tabular}{c} 
      \includegraphics[scale=0.4]{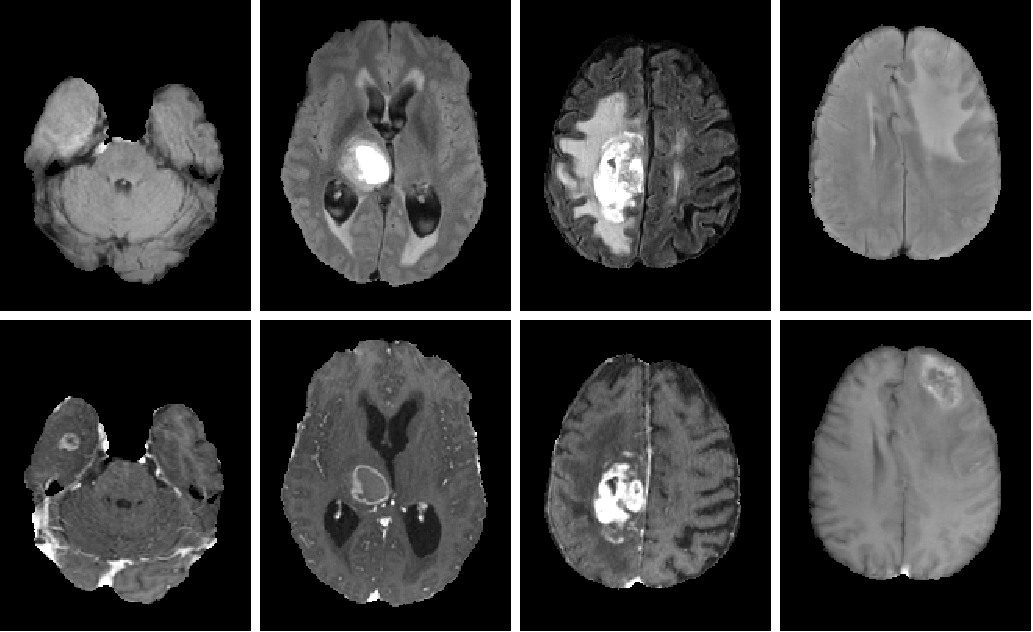}\\
    \end{tabular}
  \end{center}
  \caption{Image pairs FLAIR(top) and T1Gd(bottom) for four patients with GBM, extracted from BRATS
  dataset~\cite{menze2014multimodal, bakas2017advancing, bakas2018identifying}. Note the differences in shape, size, intensity, and contrast.}
  \label{fig:diff_images}
\end{figure}

GBM tumor segmentation consists in delineating each region that composes a lesion: necrotic core (NC), enhancing tumor (ET), and vasogenic edema (ED). For that, it is necessary to use MRI sequences that reveal the multiple visual aspects of those regions. For example, FLAIR (T2 fluid-attenuated inversion recovery) activates the edema while T1Gd (post-contrast T1-weighted) activates the enhancing tumor and shows the necrotic core as a non-activated region~\cite{osborn2015diagnostic, bakas2018identifying}. Some works show that additional T1 and T2 scans can improve the results, but not significantly compared to FLAIR and T1Gd only~\cite{zhong20202wm}. Figure \ref{fig:regions} illustrates the different visual aspects of the regions of interest in the FLAIR and T1Gd scans, showing ED, ET, and NC on the left. On the right, it shows that ED can appear with parts saturated and parts with intermediate intensity values. Such differences in appearance may also occur for ET and NC in T1Gd images. Since the selected filter bank of the first layer should activate all relevant regions, the user intervenes by drawing markers in images with such differences and choosing filters from multiple executions of FLIM with different hyperparameters. Other appearance variations in shape, intensity, contrast, and size are presented in Figure \ref{fig:diff_images}.

As mentioned in Section~\ref{sec:rw_flim}, the user draws scribbles in representative regions of the problem in a few images, and the filters are estimated from those markers by clustering. Nevertheless, the limitation in the number of desired filters requires more careful filter selection than reducing the number of filters by PCA. In MS-FLIM, the user draws markers of the same size, creating a similar amount of data on the representative regions. MS-FLIM is executed several times, first finding $N_1$ cluster centers per marker (first candidates) and subsequently clustering these first candidates into $N_2$ clusters per image. The multi-step procedure lies in the variation of $(N_1, N_2)$ under user control to explore different configurations of the patch feature space, extracting filters that are more specific and later filters that are more general. Both clustering operations use K-means~\cite{sculley2010web}.

\begin{figure}[ht]
  \begin{center}
    \begin{tabular}{c} 
      \includegraphics[scale=0.17]{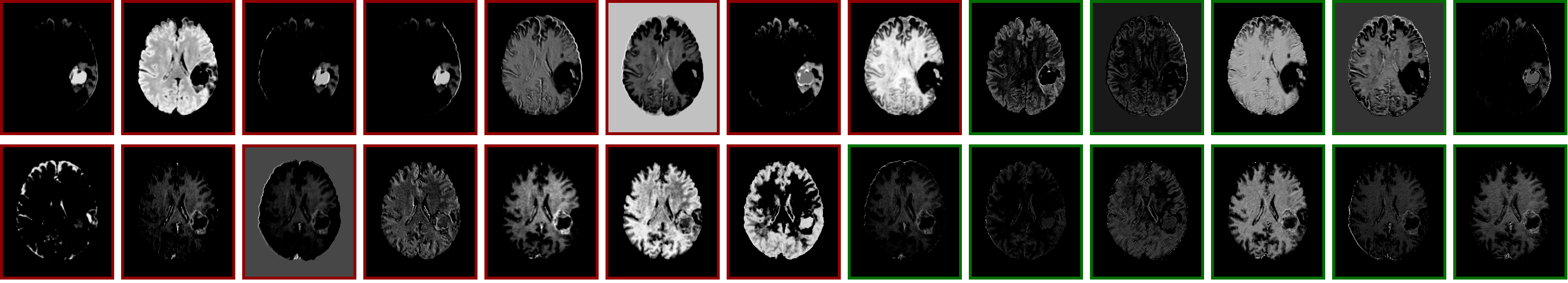}\\
      \\
      \includegraphics[scale=0.17]{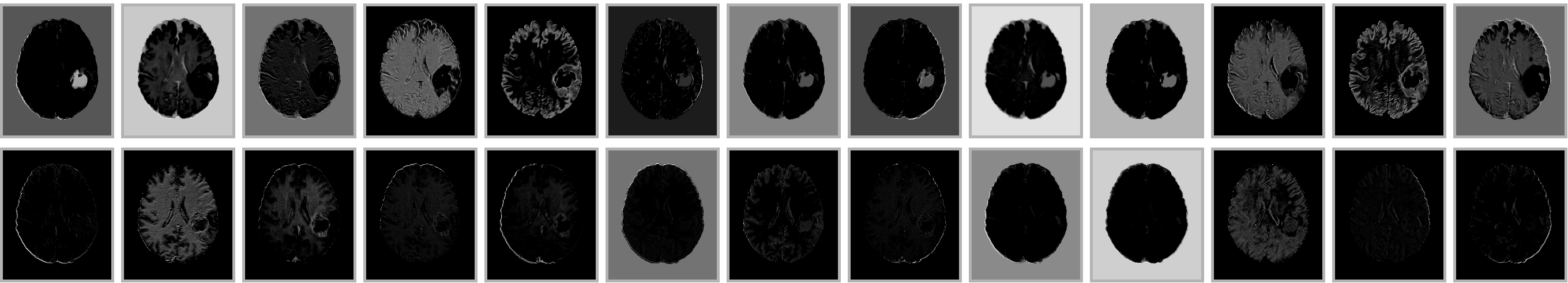}\\
    \end{tabular}
  \end{center}
  \caption{Activations obtained by MS-FLIM (top) and the standard FLIM methodology (bottom). Red-border activations were obtained with $N_1=10$ and $N_2=5$ while green-border activations were obtained with $N_1=10$ and $N_2=50$. Note the amount of relevant activations is considerably higher with MS-FLIM compared to FLIM.}
  \label{fig:features}
\end{figure}

For each execution, the user verifies which filters activate the regions of interest and selects those filters for the kernel bank of the first layer. In FLAIR, the user searches filters that activate both saturated and intermediate-intensity parts of the edema. In T1Gd, the user searches filters that activate the active tumor and the necrotic core in images where those regions appear differently. Thus, the user selects the desired number of filters in the first layer, ensuring that kernels activate all desired regions of the training images. Then one can execute the standard FLIM methodology for the rest of the encoder. Figure \ref{fig:features} shows an example of activations obtained by the standard FLIM methodology (bottom) and  MS-FLIM (top). Note the number of irrelevant filters created by FLIM compared to the representative activations obtained by MS-FLIM. Red-border activations were obtained with $N_1=10$ and $N_2=5$ while green-border activations were obtained with $N_1=10$ and $N_2=50$.

\section{Experimental Setup}
\label{sec:setup}

This section presents the experimental setup to validate MS-FLIM. We built a shallow U-shaped encoder-decoder network (Figure~\ref{fig:s_unet}), named \textit{sU-Net}, for GBM tumor segmentation into three regions: ET, NC, and Whole Tumor (WT). The literature usually reports the segmentation effectiveness for these three regions, assuming that $\rm{ED} = \rm{WT} \setminus (\rm{ET} \cup \rm{NC})$. We used DICE (DSC) to measure effectiveness. We also compared the sU-Net-based methods with two SOTA baselines on two datasets. 

\subsection{Datasets}
\label{sec:datasets}

The experiments used a private dataset containing 80 3D images of GBM (HGG) with two 3T-MRI scans (FLAIR and T1Gd) per patient. Image acquisition was volumetric with voxels of $1mm^3$. We co-registered the images, skull stripped~\cite{martins2019adaptive}. We used the private dataset to train all models and the training dataset of BRATS 2021 to verify how well the models generalize to a larger dataset, including lesions different from GBM. 

We randomly divided the private dataset into 70\% for training, 10\% for validation, and 20\% for testing. Once trained all models, we verify their ability to generalize by evaluating the best models over the BRATS 2021 training set, which consists of 1251 samples of brain tumors (HGG and LGG).

\subsection{sU-Net-based methods}
\label{sec:sNet}

\begin{figure}[h]
  \begin{center}
    \begin{tabular}{c} 
      \includegraphics[scale=0.35]{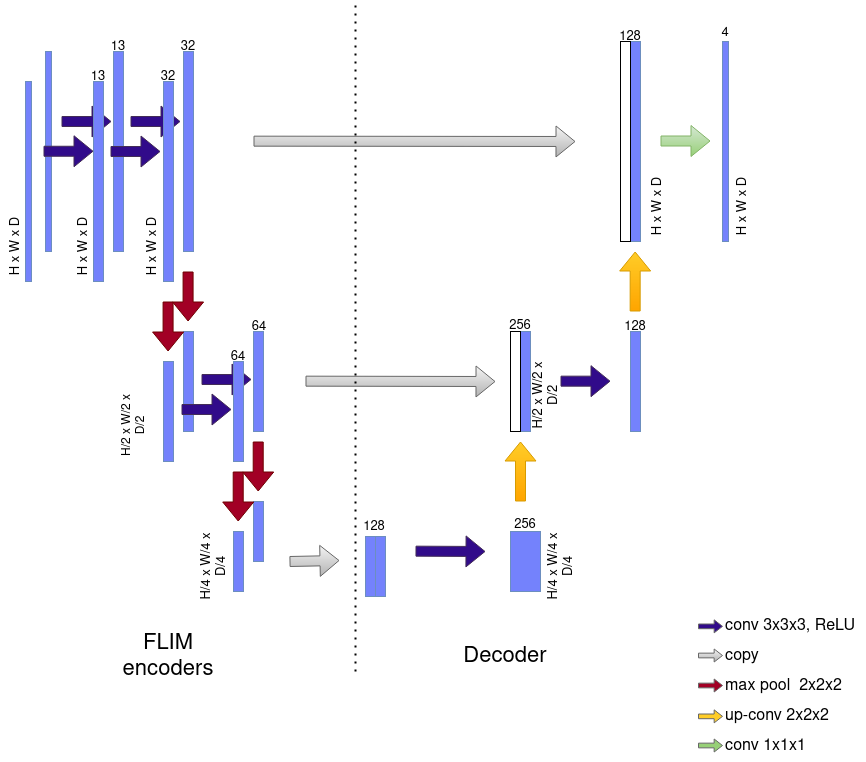}
    \end{tabular}
  \end{center}
  \caption{sU-Net architecture with two encoders, one for FLAIR and the other for T1Gd scans.}
  \label{fig:s_unet}
\end{figure}

The sU-Net architecture (Figure~\ref{fig:s_unet}) consists of two encoders, one for T1Gd and the other for FLAIR images, with three convolutional layers each. Skip connections concatenate the output feature block, before each stridden pooling operation, with the output of a decoder's layer after the transposed convolution. The skip connections provide fine-grained segmentation, and in the final layer, a convolution with kernels $1^3$ generates four channels, one for the background and one for each label (ED, ET, NC).

We compared four methods based on the sU-Net architecture: sU-Net (FBp), in which encoder and decoder are trained with backpropagation only (i.e., FBp - Fully Backpropagation); sU-Net (FLIM and PBp), in which the encoder is initialized by FLIM and only the decoder is trained by backpropagation (i.e., PBp - Partial Backpropagation); sU-Net (MS-FLIM and PBp), in which the encoder is initialized with MS-FLIM and only the decoder is trained with backpropagation; and sU-Net (MS-FLIM and FBp), in which backpropagation is used to train the decoder and fine-tune the encoder initialized by MS-FLIM. Comparing the last and previous models aims to verify how optimum was the MS-FLIM initialization.   

FLIM and MS-FLIM encoders initialization were performed from the same user-drawn markers on eight training images. Each encoder was trained with a subset of six of those eigth images (FLAIR and T1Gd); We manually selected each image for encoder training. Complete image annotation was used only for backpropagation, wherein each run, we used the exact configuration of data split, learning rate ($2.5e^{-3}$ with linear decay), loss (average of Cross-Entropy and Dice) and a total of 100 epochs. We also used ADAM optimizer and batch size equal to one.

\subsection{Baselines}
\label{sec:baselines}

DeepMedic~\footnote{\url{https://github.com/deepmedic/deepmedic} } and nnU-Net~\footnote{\url{https://github.com/MIC-DKFZ/nnUNet} } models were used as baselines. These models use 3D patches as input rather than 3D images and adopt data augmentation, normalization, and learning rate reduction, providing us with upper-bound metrics. DeepMedic was trained for 100 epochs and nnU-Net for 1000 epochs; Other configuration (batch size, optimizer, etc.) remains unchanged. The goal was to verify how close our simple sU-Net models can be to these SOTA models.

\section{Results and Discussion}
\label{sec:results}

Table \ref{tab:results_ours} shows the results using the private dataset. As expected, nnU-Net and DeepMedic performed better than the sU-Net-based models. However, the differences are comparable to the standard deviation in the case of MS-FLIM, indicating the proposed method's high quality. Furthermore, the sU-Net-based models did not adopt data augmentation.
 
 The training times of these models are also an interesting comparison: while sU-Net (FLIM and PBp) takes a few hours, DeepMedic takes several hours, and nnU-Net takes a few days. The training time of sU-Net (MS-FLIM and PBp/FBp, both) depends on the user experience and also takes a few hours due to the backpropagation and drawing markers.
 
 Among the sU-Net-based models, it is clear the improvement of sU-Net (MS-FLIM and PBp/FBp) over sU-Net (FPb) and sU-Net (FLIM and PBp). Comparing the two models based on MS-FLIM indicates that the fine-tuning of the encoder by backpropagation was unnecessary. 
 
\begin{table}[ht]
\caption{Results for our data} 
\label{tab:results_ours}
\begin{center}       
\begin{tabular}{|l|l|l|l|}
\hline
\rule[-1ex]{0pt}{3.5ex}  Models & \multicolumn{3}{|c|}{DSC}  \\
\hline
\rule[-1ex]{0pt}{3.5ex}    & ET & NC & WT   \\
\hline
\rule[-1ex]{0pt}{3.5ex}  DeepMedic & 0.777(0,056)  & 0.851(0,066) & 0.792(0,094)    \\
\hline
\rule[-1ex]{0pt}{3.5ex}  nnU-Net & \textbf{0.798(0,045)} & \textbf{0.885(0,058)} & \textbf{0.851(0,068)}   \\
\hline
\rule[-1ex]{0pt}{3.5ex}  sU-Net (FBp)& 0.665(0,166) & 0,734(0,157) & 0,721(0,104)   \\
\hline
\rule[-1ex]{0pt}{3.5ex} sU-Net (FLIM and PBp) & 0,691(0,073) & 0,733(0,072) & 0,702(0,109)  \\
\hline 
\rule[-1ex]{0pt}{3.5ex} sU-Net (MS-FLIM and PBp) & 0.746(0,052) & \textbf{0.813(0,073)} & \textbf{0.785(0,082)}  \\
\hline
\rule[-1ex]{0pt}{3.5ex} sU-Net (MS-FLIM and FBp) & \textbf{0,747(0,051)} & 0,805(0,078) & 0,780(0,096)  \\
\hline 
\end{tabular}
\end{center}
\end{table}
 
\begin{figure}[ht]
  \begin{center}
    \begin{tabular}{c} 
      \includegraphics[scale=0.3]{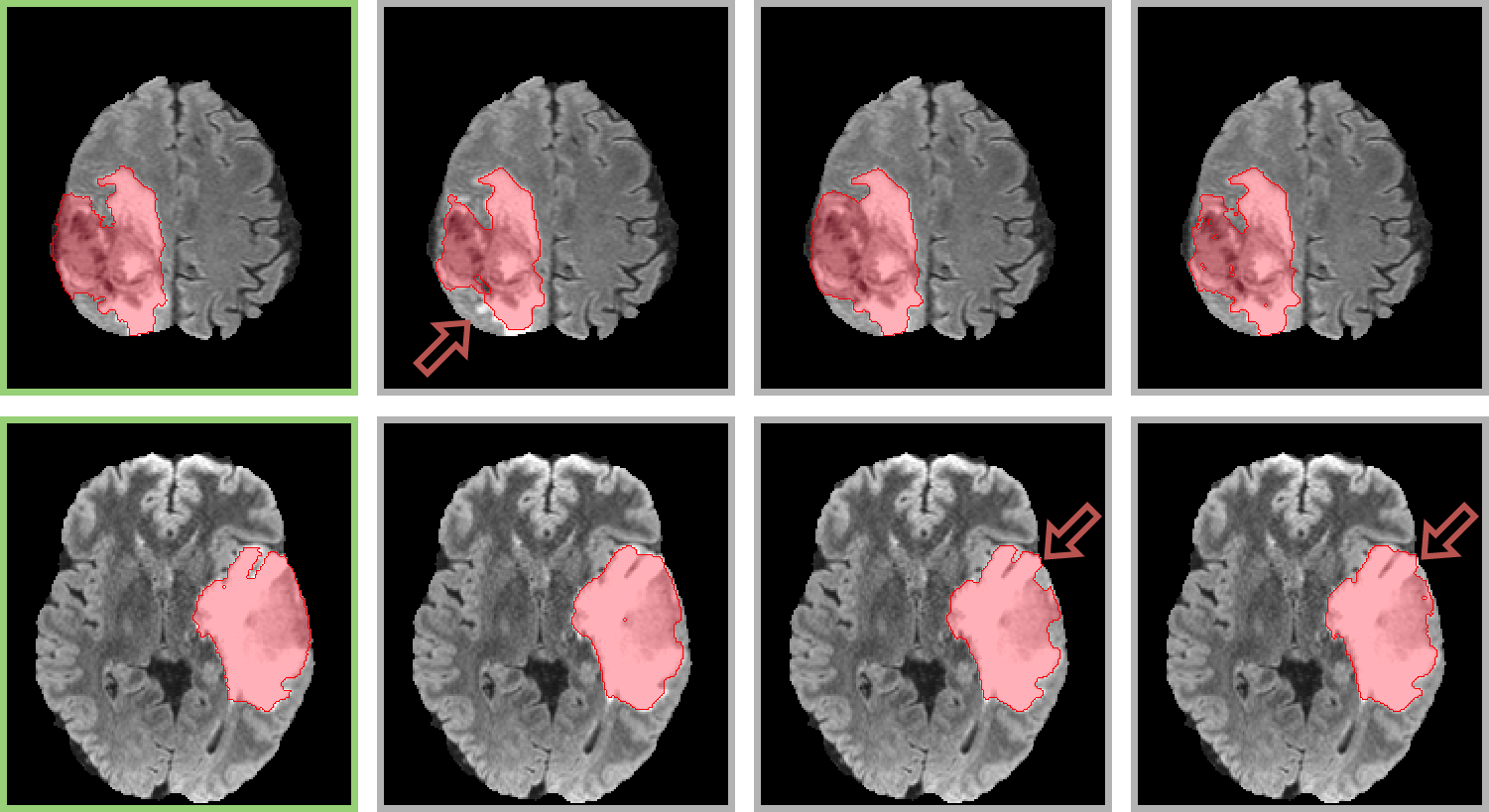}\\
    \end{tabular}
  \end{center}
  \caption{Whole tumor segmentation for two patients from private dataset, columns are from Ground Truth (GT), MS-FLIM, nnU-Net and DeepMedic.}
  \label{fig:visual_results}
\end{figure}

Figure \ref{fig:visual_results} shows qualitative results for two samples of the test set of our private dataset. The top row contains an example where the MS-FLIM segmentation performs poorly compared to nnU-Net and DeepMedic, missing a region (arrow). Alternatively, the bottom row shows a case that MS-FLIM achieved better results than nnU-net and DeepMedic. Finally, the figure illustrates that our model can achieve better results in several situations and that the main distinctions between our results and the upper-bound (SOTA deep learning models) are in the fine-grained segmentation.

As said in Section \ref{sec:setup}, we verified how well the models generalize to the BRATS dataset. Thus, Table \ref{tab:results_brats} contains the evaluation metrics for all 1251 images of the training dataset of BRATS 2021. From that, one can notice that MS-FLIM reached an average score slightly below the standard backpropagation algorithm, which indicates that MS-FLIM was more specialized on the dataset with GBM lesions, however their differences are within the standard deviations.
 
\begin{table}[ht]
\caption{Results for BRATS dataset} 
\label{tab:results_brats}
\begin{center}       
\begin{tabular}{|l|l|l|l|}
\hline
\rule[-1ex]{0pt}{3.5ex}  Models & \multicolumn{3}{|c|}{DSC}  \\
\hline
\rule[-1ex]{0pt}{3.5ex}    & ET & NC & WT  \\
\hline
\rule[-1ex]{0pt}{3.5ex}  nnU-Net & 0,756(0,230)  & 0,805(0,259) & 0,840(0,167)  \\
\hline
\rule[-1ex]{0pt}{3.5ex}  DeepMedic & 0,630(0,273) & 0,662(0,297) & 0,674(0,233)   \\
\hline
\rule[-1ex]{0pt}{3.5ex}  sU-Net (FBp)& 0,554(0,223) & 0,629(0,257) & 0,722(0,182)  \\
\hline
\rule[-1ex]{0pt}{3.5ex}  sU-Net (MS-FLIM and PBp) & 0,561(0,228) & 0,603(0,264) & 0,708(0,184)   \\
\hline
\rule[-1ex]{0pt}{3.5ex}  sU-Net (MS-FLIM and FBp) & 0,563(0,234) & 0,601(0,267) & 0,704(0,190)  \\
\hline 
\end{tabular}
\end{center}
\end{table}

\section{Conclusion}
\label{sec:conclusion}

This work has presented an improvement over the recent FLIM methodology, which initializes the filters of an encoder from user-drawn markers on problem-related regions of a few images. The new method, named MS-FLIM, additionally involves the user in filter estimation and selection to obtain the most relevant filters for a given convolutional layer from multiple executions of the FLIM algorithm. We used MS-FLIM in the first convolutional layer only, and the results on our private dataset of GBM tumor segmentation already show improvements over the same architecture trained with FLIM and backpropagation only. A simple network with an encoder initialized by MS-FLIM could achieve effectiveness within the standard deviations of the SOTA models used as references. 

Since MS-FLIM required markers on only eight training images, we intend to investigate its extension to initialize decoders, eliminating the use of backpropagation with many fully annotated images (i.e., considerably reducing user involvement in data annotation). We also intend to evaluate MS-FLIM with user interaction in more layers, other types of decoders, and for deeper models. 

\acknowledgments 
 
The authors thank FAPESP (2014/12236-1), CNPq (303808/2018-7 and 130418/2021-8) for financial support.

\bibliography{references} 
\bibliographystyle{spiebib} 

\end{document}